\documentclass[10pt,journal,compsoc]{IEEEtran}

\ifCLASSOPTIONcompsoc
  \usepackage[nocompress]{cite}
\else
  \usepackage{cite}
\fi

\ifCLASSINFOpdf
\else
\fi

\usepackage[english]{babel} % English language hyphenation

\usepackage{graphicx} % Required for including images

\usepackage{epstopdf}
\usepackage[numbers,sort&compress]{natbib}

\usepackage{subfigure}
\usepackage{graphicx}
\usepackage{wrapfig} % Allows in-line images

\usepackage{booktabs} % Required for better horizontal rules in tables

\usepackage{algorithm}  
\usepackage{algpseudocode}  
\usepackage{amsmath}
\usepackage{amssymb}
\usepackage{mathrsfs}
\usepackage{CJK}
\usepackage{indentfirst} 
\usepackage{enumitem} % Required for list customisation
\setlist{noitemsep} % No spacing between list items

\makeatletter
\renewcommand\@biblabel[1]{\textbf{#1.}} % Remove the square brackets from each bibliography item ('[1]' to '1.')
\makeatother

\usepackage{mathrsfs}
\usepackage{amsfonts}
\usepackage{booktabs}
\usepackage{enumitem}
\usepackage{amsthm}

\begin{document}

\title{Representation Learning of \\ Reconstructed Graphs Using Random \\ Walk Graph Convolutional Network}

\author{Xing~Li,
        Wei~Wei,
        Xiangnan~Feng,
		Zhiming~Zheng
\IEEEcompsocitemizethanks{\IEEEcompsocthanksitem W. Wei was with the School of Mathematical Science, Beihang University, Beijing, China, Key Laboratory of Mathematics Informatics Behavioral Semantics, Ministry of Education, China, Peng Cheng Laboratory, Shenzhen, Guangdong, China and Beijing Advanced Innovation Center for Big Data and Brain Computing, Beihang University, Beijing, China. \protect\\

E-mail: see weiw@buaa.edu.cn
\IEEEcompsocthanksitem X. Li, X. Feng and Z. Zheng were with the School of Mathematical Science, Beihang University, Beijing, China, Key Laboratory of Mathematics Informatics Behavioral Semantics, Ministry of Education, China and Beijing Advanced Innovation Center for Big Data and Brain Computing, Beihang University, Beijing, China.}% <-this % stops an unwanted space
\thanks{ }}

\IEEEtitleabstractindextext{%
\begin{abstract}
Graphs are often used to organize data because of their simple topological structure, and therefore play a key role in machine learning. And it turns out that the low-dimensional embedded representation obtained by graph representation learning are extremely useful in various typical tasks, such as node classification, content recommendation and link prediction. However, the existing methods mostly start from the microstructure (i.e., the edges) in the graph, ignoring the mesoscopic structure (high-order local structure). Here, we propose wGCN -- a novel framework that utilizes random walk to obtain the node-specific mesoscopic structures of the graph, and utilizes these mesoscopic structures to reconstruct the graph And organize the characteristic information of the nodes. Our method can effectively generate node embeddings for previously unseen data, which has been proven in a series of experiments conducted on citation networks and social networks (our method has advantages over baseline methods). We believe that combining high-order local structural information can more efficiently explore the potential of the network, which will greatly improve the learning efficiency of graph neural network and promote the establishment of new learning models.
\end{abstract}

}

\maketitle

\IEEEdisplaynontitleabstractindextext

\IEEEpeerreviewmaketitle

\IEEEraisesectionheading{\section{Introduction}\label{sec:introduction}}

\IEEEPARstart{G}{raph} is a flexible data structure, which can store data and reflect the underlying topological relationship between data. Because of this, graph structure is widely used in various fields, including social networks \cite{1}, biological protein-protein networks \cite{2}, drug molecule graphs \cite{3}, knowledge network \cite{4}, etc. For example, in a drug molecule graph, a node can represent an atom, and nodes are connected with edges indicating that there are chemical bonds between atoms. By this way, people can easily store the data information in the network and access the relational knowledge about the interactive entities from this structured knowledge base at any time.

The traditional methods of extracting information from graphs depend on the statistics (degree, clustering coefficient \cite{5}, centrality \cite{6,7}, etc.) of graphs, or kernel functions \cite{8} (or other characteristic functions) which are carefully designed. However, with the development of information technology, the amount of data increases rapidly, which makes the graph network more and more complex. Therefore, the traditional manual feature extraction methods becomes expensive, time-consuming and unreliable -- can not extract valid information from complicated organizations.

In this case, representation learning has played a key role to efficiently extracts information in the graph. The graph representation learning method is a technical method for learning graph structure data, which hopes to transform complex original data into easy-to-process low-dimensional vector representations. In essence, the graph representation learning method is to learn a function, and this function maps the input graph or node to a point in the low-dimensional vector space. Compared with traditional methods, the representation learning method treats the problem of capturing graph information as part of the learning task itself, rather than just as a preprocessing link. In fact, the representation learning method uses data-driven methods to obtain features, avoiding the trouble of traditional manual feature extraction.

The goal of representation learning is not simply to obtain results directly, but to obtain an efficient representation of the original data. In other words, the choice of representation usually depends on subsequent learning tasks, and a good representation should make the learning of downstream tasks easier. In recent years, representation learning on graphs is very popular, and many good results have been obtained. Belkin et al. \cite{9} propose Laplacian feature map in 2002, which is one of the earliest and most famous representation learning methods. Then a large number of methods are proposed, and the representative ones such as Grarep (Cao et al. \cite{10}), Deepwalk (Perozzi et al. \cite{11}), node2vec (Grover et al. \cite{12}), GNN (Scarselli et al. \cite{13}), GCN (Kipf et al. \cite{14}), etc.

However, on the one hand, these methods are based on the binary relationship (i.e., edges) in the graph, and can not leverage the local structure of the graph; on the other hand, due to the sparseness of the edges in the graph, many methods are Encountered difficulties in generalization in many cases. Thus, our method is proposed in order to leverage the high-order connection patterns that are essential for understanding the control and regulation of the basic structure of complex network systems, and to alleviate the problem of edge sparsity in the graph.
\vspace{2ex}

\noindent\textbf{Present work.} This paper develops a new framework -- wGCN, a controllable and supervised representation learning method. wGCN can be regarded as a two-stage task: in the first stage, according to the original graph structure data, we execute a random walk algorithm on the graph, and then use the generated walk to redistribute the weight of the graph network to obtain the reconstructed graph; in the second stage, we connect the reconstructed image to the graph convolutional neural network, and combine the original features and labels for training. This new framework can improve the accuracy of prediction tasks while spending a little more time. We prove that it has the same level of time complexity as the graph convolutional neural network, and conduct a large number of experiments on a variety of actual datasets, all of which have obtained better test results than the baselines.

The rest of this article is organized as follows. In Section 2, the related work in the past is summarized, and in Section 3, our representation learning method wGCN is introduced in detail. Our experiment will be introduced in Section 4, and the results will be given in Section 5. In Section 6, conclusions and future work will be discussed.

\section{Related Work}

This part will focus on previous work closely related to wGCN. Specifically, we first introduce random walk, and then introduce some classic graph representation learning methods, which inspires our methods.

\subsection{Random Walk}

Random walk refers to the behavior that can not predict future development steps and directions based on past performance. The core concept is that the conserved quantities of any irregular walker correspond to a law of diffusion and transportation, which is the ideal mathematical state of Brownian motion.

\begin{figure}
\center
\includegraphics[width=\linewidth]{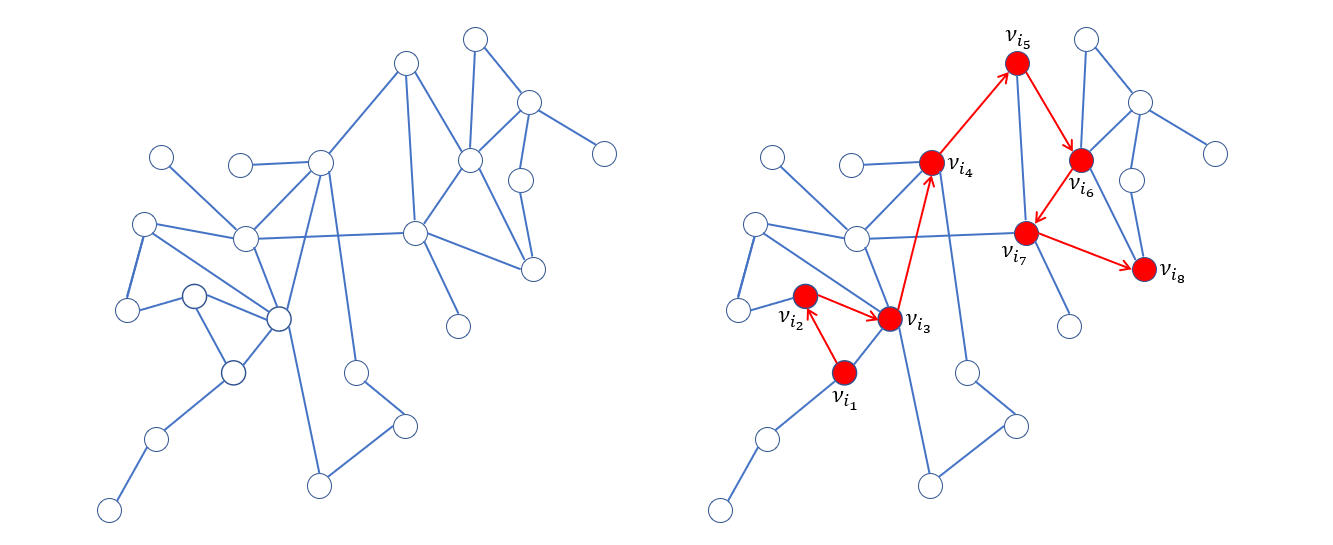}
\caption{The image on the left is the original graph, and the image on the right shows a random walk of length 8 (8 nodes, marked in red). The red arrow shows the walking order, and this random walk is recorded as $v_{i_1}v_{i_2}...v_{i_8}.$}
\label{fig:1} 
\end{figure}

The random walk on the graph refers to starting from one or a series of nodes and moving between nodes in the graph according to specific rules. For example, a walker randomly selects a vertex to start, walks to the neighbor node of this vertex with probability $p \ (0<p\leq1)$, and jumps to any node in the graph with probability $1-p$, which is called a jump-turn probability. Each walk will result in a probability distribution reflecting the node to be visited, which is used as the input of the next step of the random walk. And the above process is iterated continuously. When certain conditions are met, the results of the iteration will tend to converge, resulting in a stable probability distribution. Fig.1 gives a specific random walk on graph. Random walk is widely used in the field of information retrieval. The well-known PageRank \cite{15} algorithm is a classic application of random walk.

\subsection{Representation Learning Method}

The graph representation learning method is a technical method for learning graph structure data. It hopes to transform complex raw data into a low-dimensional representation that is convenient to develop and process by machine learning. To a certain extent, it can also be regarded as a method of dimensionality reduction. According to whether neural network is used, graph representation learning methods can be divided into two categories: i) direct coding methods that do not use neural networks; ii) neural network-based coding methods.
\vspace{2ex}

\noindent\textbf{Direct coding methods.} Early learning node representation methods are concentrated in the matrix factorization framework. Belkin et al. present Laplacian eigenmaps method, which is a encoder-decoder framework measured by Euclidean distance in the coding space \cite{9}. Following the Laplacian eigenmaps method, Ahmed et al. \cite{16} and Cao et al. \cite{10} propose Graph Factorization (GF) and GraRep separately, whose main difference is the way the basic matrix is used. The original adjacency matrix of graph is used in GF and GraRep is based on various powers high order relationship of the adjacency matrix. And Mingdong er al. present High Order Proximity preserved Embedding (HOPE), which can preserve the asymmetric transitivity of the directed graph \cite{17}. 

\begin{figure}
\subfigure[]{
\begin{minipage}[t]{0.2\linewidth}
\centering
\includegraphics[width=1.2in]{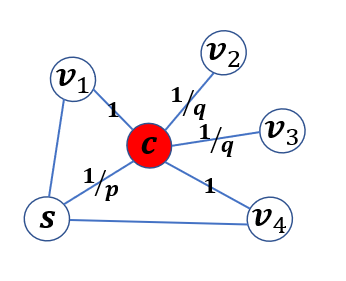}
\end{minipage}
}%
\subfigure[]{
\begin{minipage}[t]{0.85\linewidth}
\centering
\includegraphics[width=2.2in]{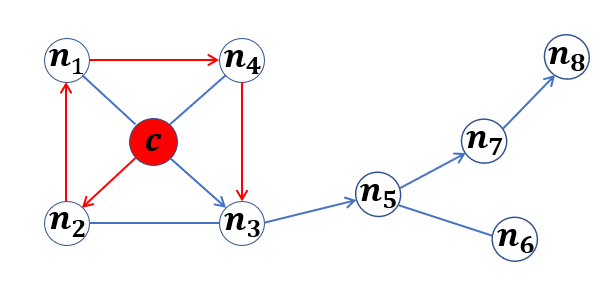}
\end{minipage}
}%
\caption{\textbf{Explanation of node2vec.} In subfigure (a), node $s$ is the source node of a random walk, and node $c$ is the current node. Hyperparameters $p$ and $q$ control the walking probability of the next step (from node $c$ to its neighbors). As marked in subfigure (a), the probability of returning node $s$ from node $c$ is $1/p$; the probability from node $c$ to node $v_1$ or $v_4$ is $1$ as $v_1$ and $v_4$ are also neighbors of node $s$; and the probability from node $c$ to node $v_2$ or $v_3$ is $1/q$ as $v_2$ and $v_3$ are the second-order neighbors of node $c$. Thus, when $p$ decreases, walking tends to return to the source node (i.e., "more local"); when $q$ decreases, walking tends to move away from the source node (i.e., "more global"). The consequence is shown in subfigure (b), red arrows show the walking which is "more local", and blue arrows show the walking which is "more global".}
\label{fig:2}
\end{figure}

On the other hand, there are also some classic methods based on random walk instead of matrix factorization, thus becoming more flexible. Here, two representative methods are DeepWalk \cite{11} and node2vec \cite{12}. DeepWalk uses random walk to disassemble the graph which is nonlinear structure into multiple linear node sequences, then the node sequences treated as "sentences" (the nodes are treated as "words") are processed by SkipGram \cite{18}. As for node2vec, it allows an adjustable random walk on the graph. In particular, node2vec creatively uses two hyperparameters $p$ and $q$ to control the random walk "more local" or "more global", in other words, depth-first search or breadth-first search (relative to the starting node, see Fig.2).

\vspace{2ex}

\noindent\textbf{Neural network-based coding methods.} The above direct encoding methods independently generate a representation vector for each node, which results in no shared parameters between nodes, high computational complexity, and underutilized node attribute information. Considering to solve these problems, many graph neural network methods have been proposed in recent years. Scarselli et al. \cite{13} present the Graph Neural Network (GNN) model which can implement a function that maps the graph and one of its nodes to Euclidean space. And Inspired by the success of Convolutional Neural Network in image processing (Convolutional Neural Network extremely reduces the number of parameters by using convolution kernels to gather the information of local pixels on the image. However, CNN has encountered difficulties on the graph, due to the irregularity of the graph, that is, the number of neighbors is uncertain), Kipf et al. \cite{14} propose the well-known Graph Convolutional Network (GCN) method. The GCN method cleverly applies the convolution operation to the graph structure, which means that the information of neighbor nodes is aggregated on the irregular graph structure.
\vspace{4ex}

\section{Method}

We first perform random walk operation on the original graph, and then use the obtained "walks" to reconstruct the graph network. After that, the graph convolution operation is performed on the obtained reconstructed graph to obtain the representation vectors of the nodes. We believe that these approaches can combine the nodes of the graph at multiple levels to obtain a more informative representation. Finally, the obtained representation vectors are sent to the downstream classifier (such as knn, mlp and so on) to complete the node classification task.

Next, we will introduce the technical details of our method. First, the random walk and the reconstructed graph is introduced in detail in Section 3.1; then, the wGCN embedding algorithm to generate embeddings for nodes is described in Section 3.2; finally in Section 3.3, we give an analysis of the complexity of the algorithm and prove it at the same time.

\subsection{Reconstructed Graph}

We will explain reconstructed graph utilizing random walks in detail in this section. For the convenience of explanation, some commonly used symbols are given below:

Formally, let graph $G=(V,E)$, where $V$ is the set of the nodes in network, $|V|$ represents the number of nodes, and $E \subseteq (V \times V)$ is the set of the edges. Given a labeled network with node feature information $G=(V,E,X,Y)$, where $X \in R^{(N \times T)}$ ($N$ is the number of the nodes in network, $T$ is the feature dimension) is the feature information matrix and $Y \in R^{(N \times L)}$ ($L$ is the feature dimension) is the label information matrix, our goal is to use the labels of some of the nodes for training, and generate a vector representation matrix $Z$ of the nodes.

Then, we give:

\noindent\textbf{Definition 1:} Given a graph $G=(V,E,X,Y)$ ($V=\{v_1, v_2, ..., v_{|V|}\}$) and initial node $v_{i_1}$, a random walk of length $n$ rooted at $v_{i_1}$ is denoted as $W_{v_{i_1}}=\{(v_{i_1},v_{i_2}),(v_{i_2},v_{i_3}),...,{(v_{i_{n-1}},v_{i_n})}\}$, where $1 \le i_1,i_2,...,i_n \le |V|$ and the two nodes in the same parenthesis are neighbors (i.e., there is an edge connection between $v_{i_j}$ and $v_{i_{j+1}}$). Or for convenience, $W_{v_{i_1}}=\{v_{i_1}v_{i_2}v_{i_3}...v_{i_{n-1}}v_{i_n}\}$.

\begin{figure*}
\begin{center}
\includegraphics[width=0.9\linewidth]{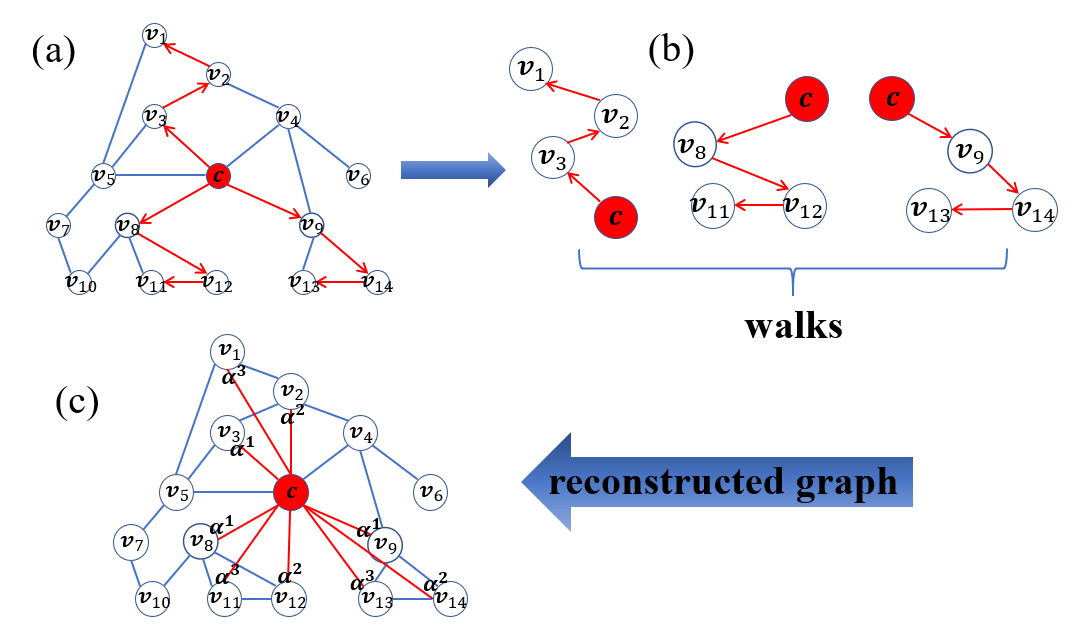}
\end{center}
\caption{\textbf{Example of reconstructing graph using random walk.} Subfigure (a) is the original graph, in which the node $c$ is the source node marked in red. And the "walks" are marked with red lines and extracted in subfigure (b). Subfigure (c) is the reconstructed graph, where the weights $\alpha^k,k=1,2,3$ is used to weight the original k-hop nodes away from node $c$.}
\label{fig:1} 
\end{figure*}

As shown in the Fig.3, we first perform random walk operation on the graph (subfigure (a)) to obtain "walks" (subfigure (b)). After that, we can use the obtained "walks" to reconstruct the graph network. A node that appears in the same walk with node $c$ is considered to be related to node $c$, and this node is connected to node $c$ in the reconstructed graph (the red lines in subfigure (c)). And we assign different weights to distinguish the distance between the nodes and node $c$ in "walks". Thus in the reconstructed graph, the weight $w_{ij}$ for nodes $i,j$ can be given by the following formula:

\begin{equation}
w_{ij}=a_{ij}+\sum_{k \in \{the ~ spacing ~ of ~ i,j ~ in ~ "walks"\}}\alpha^k,
\end{equation}

\noindent where $a_{ij}$ indicates the original adjacency matrix, which is defined as follows:

\begin{equation}
a_{ij}=\begin{cases}
0,& \text{if nodes $i,j$ are not connected in the original graph}\\
1,& \text{if nodes $i,j$ are connected in the original graph}
\end{cases},
\end{equation}

\noindent and $\alpha$ is a parameter to be determined indicating the decay speed with distance, satisfying $0<\alpha<1$. And k is the exponent, whose values are the distance between nodes $i,j$ in the "walks". For example, node $c$ and node $v_{{}_{12}}$ are not connected in the original graph (subfigure (a)), thus $a_{cv_{{}_{12}}}=0$; and node $c$ and node $v_{{}_{12}}$ have a distance of $2$ in "walks" (subfigure (b)), thus $k=2$; in summary:

\begin{eqnarray}
w_{cv_{{}_{12}}}&=&a_{cv_{{}_{12}}}+\sum_{k \in \{2\}}\alpha^k \nonumber     \\
&=&0+\alpha^2    \\
&=&\alpha^2 \nonumber.
\end{eqnarray}

\noindent In the next section, we give the complete algorithm.

\subsection{Embedding Algorithm}

\begin{algorithm}
	\caption{\textbf{:} wGCN embedding algorithm.}  
	\label{alg:Framwork}  
	\begin{algorithmic}[1]  
    	\Require  
      		Graph $(V,E)$;
			Adjacency matrix $A$;
			The number of walks from each node $T$;
			Walks length $L$;
			Reconstructed scale parameter $\lambda$;
			Node feature matrix $X$;
			The number of graph convolutional neural network layers $H$; 
    	\Ensure  
     	 	Representation matrix $Z$;
		\State Initialize the random walk matrix $\mathcal{A}$ as zero;
		\For{$t=1...T$}
			\For{$v \in V$}
				\State $W_{v}^t=RandomWalk(G,v,L)$
			\EndFor
		\EndFor
		\For{$t=1...T$}
			\For{$v \in V$}
				\For{$l=2...L$}
					\State $\alpha_{vW_{v}^t[l]}=\alpha_{vW_{v}^t[l]}+\alpha^{l-1}$;
				\EndFor
			\EndFor
		\EndFor
		\State $\tilde{A}=A+ \lambda \cdot \mathcal{A}$
    	\State $h_v^{0}=x_v, \forall v \in V$;
		\For{$k=1...H$}
			\For{$v \in V$}
				\State $h_v^{k} \leftarrow f(h_v^{k-1},h_u^{k-1}),u \in N_{\tilde{A}}(v)$;
			\EndFor
		\EndFor
    	\State $z_v=h_v^{H}, \forall v \in V$;
    	\State \textbf{return} $Z$ (whose column vectors are $z_v, v \in V$); 
	\end{algorithmic}  
\end{algorithm}

After entering the required data, we first initialize the random walk matrix $\mathcal{A}$ to a zero matrix in line 1. The random walk matrix $\mathcal{A}$ refers to the matrix generated by random walk, whose element $\alpha_{uv}$ indicates the weight attached to the nodes $u,v$ by the "walks". And in lines 2-14, the reconstructed graph matrix is built as described in section 3.1:

\noindent 1. Generate $T$ "walks" of length $L$ at each node ($RandomWalk(\cdot)$) in lines 2-5;

\noindent 2. Update the elements of the random walk matrix $\mathcal{A}$ in lines 7-13;

\noindent 3. Obtain the reconstructed graph matrix $\tilde{A}$ in line 14.

Note that the "walk" returned by $RandomWalk(\cdot)$ is a node list. For example, $W_{v_{1}}^1=RandomWalk(G,v_{1},L)=[v_{1}v_{2}...v_{L}]$, in which $v_{i+1}$ is the neighbor of $v_{i}$, $1 \leq i \leq L-1$, and $W_{v_{1}}^1[2]=v_{2}$. In addition, the adjacency matrix $A$ and the random walk matrix $\mathcal{A}$ are mixed and normalized to obtain the reconstructed graph matrix $\tilde{A}$ in line 14, where $\lambda$ is the mixing ratio coefficient. And $Normalized(\cdot)$ is a symmetric normalization function:

\begin{equation}
Normalized(X)=D^{-\frac{1}{2}}(I+X)D^{-\frac{1}{2}},
\end{equation}

\noindent where $X$ is a square matrix, $I$ is the identity matrix and $D$ is a diagonal matrix, satisfying $D_{ii}=\sum_{j}X_{ij}$.

Then the graph convolution operation is performed. The number of graph convolutional neural network layers is specified by users in advance. And the initial representation of all nodes is expressed as: $h_v^{0}=x_v, \forall v \in V$, in line 15; In lines 16-20, we perform a graph convolution operation based on reconstructed graph, in the formula $h_v^{k} \leftarrow f(h_v^{k-1},h_u^{k-1}),u \in N_{\tilde{A}}(v)$, $f(\cdot)$ represents a weighted nonlinear aggregation function, whose purpose is to reorganize the information of the target node and its neighbors. Formally,

\begin{equation}
h_v^{k}=\sigma(\sum_{u \in N_M(v) \cup \{v\}}\tilde{a_{vu}} \cdot h_u^{k-1} \cdot W_k),
\end{equation}

\noindent where $h_v^{k}$ is the hidden representation of node $v$ in the $k$-th layer; $\tilde{a_{vu}}$ is the number on the $v$-row and $u$-column of the reconstructed graph matrix $\tilde{A}$, indicating the closeness between nodes of $v$ and $u$; $W_k$ is the parameter matrix to be trained of layer $k$; $N_{\tilde{A}}(v)$ is the neighborhood nodes set of node $v$ in the reconstructed graph; $\sigma( \cdot )$ represents for ReLU function:

\begin{equation}
f(x)=\text{max}(0,x).
\end{equation}

Then, the final representation vector $z_v$ of node $v$ is obtained. And the representation vectors can be sent to the downstream classifier (such as softmax classifier) to obtain the predicted category vectors $\hat{y_{v}},v \in V$.

If softmax classifier is chosen (the form is as follows),

\begin{equation}
\text{Softmax}(z)_{i}=\frac{\text{exp}(z_{i})}{\sum_{j=1}^d\text{exp}(z_{j})},~~~ i=1...d,
\end{equation}

\noindent where $z$ is the representation vector, Softmax$(z)_{i}$ is the $i$-th component of the vector Softmax$(z)$, and $d$ is the dimension of the representation vector $z$,

then the cross entropy function can be used as the loss function to train the parameters of our model:

\begin{equation}
loss = \sum_v(y_v \cdot log (\hat{y_{v}}) + (1-y_v) \cdot log (1-\hat{y_{v}})),~~~v \in trainset,
\end{equation}

\noindent where $y_v$ is the true label of the node $v$.

\subsection{Complexity Analysis}

Our method is based on GCN. And from the related work of Kipf et al. \cite{14}, we know that the computational complexity of the original GCN based on the following formula is $\mathcal{O}(|\mathcal{E}|CHF)$, where $\mathcal{E}$ is the edge set of the graph:

\begin{equation}
Z = f(X,A) = softmax \left( \hat{A} ~ {\rm ReLU}\left(\hat{A}XW^{(0)}\right) W^{(1)}\right),
\end{equation}

where $A$ is the adjacency matrix and $X$ is the feature matrix. And $\hat{A}$ is the normalized processing matrix of the adjacency matrix $A$. $W^{(0)} \in \mathbb{R}^{C \times H}$ is an input-to-hidden weight matrix and $W^{(1)} \in \mathbb{R}^{H \times F}$ is a hidden-to-output weight matrix, where $C$ is input channels, $H$ is the number of feature maps in the hidden layer and $F$ is the number of feature maps in the output layer \cite{14}. 

The time of our algorithm is mainly consumed in the training phase of the neural network. Thus, we will prove that the calculation complexity of our method in the training phase of the neural network is also $\mathcal{O}(|\mathcal{E}|CHF)$, while keeping the number of hidden layers unchanged.

\begin{proof}

Let $L$ be the "walks" length in random walk, $T$ denotes the number of "walks" from each node, and $A, \mathcal{A}, \tilde{A}$ denotes the original adjacency matrix, reconstructed graph matrix and random walk matrix respectively. We compare the number of non-zero elements in $A$ and $\tilde{A}$.

Since each node has $T$ "walks" of length $L$, then the maximum number of non-zero elements per node in the random walk matrix $\mathcal{A}$ is $T(L-1)$ (in this case, the "walks" guided by the starting node have no duplicate nodes except the starting node itself).

So $\tilde{A}$ has at most $|V|T(L-1)$ non-zero elements more than $A$ ($|V|$ is the number of nodes). And in experiments, $L$ is set to 5 and $T$ is generally set to $2|\mathcal{E}|/|V|$ ($2|\mathcal{E}|/|V|$ is the average degree of the graph), thus $T(L-1) \ll \mathcal{E}$. Therefore, the computational complexity satisfies:

\begin{eqnarray}
&&\mathcal{O}((|\mathcal{E}|+|V|T(L-1))CHF) \nonumber \\
&=&\mathcal{O}((|\mathcal{E}|+2|V|(|\mathcal{E}|/|V|)(5-1)CHF) \nonumber     \\
&=&\mathcal{O}(9|\mathcal{E}|CHF)     \\
&=&\mathcal{O}(|\mathcal{E}|CHF) \nonumber.
\end{eqnarray}

\end{proof}

\section{Experiments and Result}

In section 4.1, we introduce the datasets used in the experiment, and the specific settings of the experiment are described in section 4.2. In section 4.3, a wide variety of baselines and previous approaches are introduced, and the results are shown in section 4.4.

\subsection{Datasets}

\begin{table*} % [h] forces the table to be output where it is defined in the code (it suppresses floating)
	\renewcommand{\tablename}
	\caption{\centering{\textbf{Table 1: Summary of the datasets.}}}
	\vspace{2ex}
	\renewcommand{\arraystretch}{1.4}
	\setlength\tabcolsep{15pt}
	\centering
	\label{tab:1}
	\begin{tabular}{l l c c c c c}
		\toprule
		Datasets & Types & Undirected & Nodes & Edges & Features & Classes \\
		\midrule
		Cora & Citation & yes & 2708 & 5429 & 1433 & 7 \\
		Citeseer & Citation & yes & 3327 & 4732 & 3703 & 6 \\
		Pubmed & Citation & yes & 19717 & 44338 & 500 & 3 \\
		107Ego(F) & Social & yes & 1045 & 53498 & 576 & 9 \\
		414Ego(F) & Social & yes & 159 & 3386 & 105 & 7 \\
		1684Ego(F) & Social & yes & 792 & 28048 & 319 & 16 \\
		1912Ego(F) & Social & yes & 755 & 60050 & 480 & 46 \\
		2106Ego(G) & Social & no & 2457 & 174309 & 2094 & 2 \\
		3600Ego(G) & Social & no & 2356 & 582827 & 995 & 3 \\
		5249Ego(G) & Social & no & 826 & 435569 & 2627 & 2 \\
		7461Ego(G) & Social & no & 997 & 1270141 & 2882 & 2 \\
		\bottomrule
	\end{tabular}
\end{table*}

\noindent\textbf{Citation Network Datasets: Citeseer, Cora, and Pubmed.} In these three standard citation network datasets, nodes represent documents and edges (undirected) represent citation links. The three citation network datasets contain a sparse feature vector for each document and each document has a category label \cite{14}. As shown in the Table 1, the Cora, Citeseer and Pubmed datasets contain 1433 features, 3703 features and 500 features per node respectively, and the number of label categories are 7, 6 and 3 respectively.

\noindent\textbf{Social Network Datasets: Ego-Facebook and Ego-Gplus.} For Ego-Facebook, this dataset consists of 'circles' (or 'friends lists') from Facebook \cite{19}. There are many subsets of the Ego-Facebook dataset. Take '107Ego(F)' as an example. This dataset is a network with the node '107' as the core, where the nodes represent users and the edges (undirected) represent interactions between users. And each user has a feature attribute vector and a category label. As for Ego-Gplus, it is similar to Ego-Facebook, except that the data comes from Google$+$ and the edges are directed \cite{19}. We choose the suitable subsets of Ego-Facebook and Ego-Gplus for experiments (to facilitate the distinction, (F) represents a subset of Ego-Facebook and (G) represents a subset of Ego-Gplus). And after preprocessing, the data whose information has been lost is removed.

\subsection{Experimental Setup}

\begin{table*} % [h] forces the table to be output where it is defined in the code (it suppresses floating)
	\renewcommand{\tablename}
	\caption{\centering{\textbf{Table 2: Parameter setting on social network datasets.}}}
	\vspace{2ex}
	\renewcommand{\arraystretch}{1.4}
	\setlength\tabcolsep{18pt}
	\centering
	\begin{tabular}{l c c c c c c}
		\toprule
		Datasets & $W_1$ & $W_2$ & $W_3$ & $T$ & $\alpha$ & $\lambda$ \\
		\midrule
		107Ego(F) & $\mathbb{R}^{576\times128}$ & $\mathbb{R}^{128\times32}$ & $\mathbb{R}^{32\times9}$ & 40 & 0.8 & 0.8 \\
		414Ego(F) & $\mathbb{R}^{105\times36}$ & $\mathbb{R}^{36\times12}$ & $\mathbb{R}^{12\times4}$ & 24 & 0.8 & 0.7 \\
		1684Ego(F) & $\mathbb{R}^{319\times128}$ & $\mathbb{R}^{128\times40}$ & $\mathbb{R}^{40\times13}$ & 35 & 0.5 & 0.9 \\
		1912Ego(F) & $\mathbb{R}^{480\times160}$ & $\mathbb{R}^{160\times60}$ & $\mathbb{R}^{60\times19}$ & 80 & 0.8 & 0.9 \\
		2106Ego(G) & $\mathbb{R}^{2094\times128}$ & $\mathbb{R}^{128\times16}$ & $\mathbb{R}^{16\times2}$ & 30 & 0.8 & 0.8 \\
		3600Ego(G) & $\mathbb{R}^{995\times128}$ & $\mathbb{R}^{128\times16}$ & $\mathbb{R}^{16\times3}$ & 200 & 0.5 & 0.9 \\
		5249Ego(G) & $\mathbb{R}^{2627\times256}$ & $\mathbb{R}^{256\times16}$ & $\mathbb{R}^{16\times2}$ & 50 & 0.8 & 0.9 \\
		7461Ego(G) & $\mathbb{R}^{2882\times256}$ & $\mathbb{R}^{256\times16}$ & $\mathbb{R}^{16\times2}$ & 30 & 0.8 & 0.7 \\
		\bottomrule
	\end{tabular}
\end{table*}

\noindent\textbf{Citation Network Datasets.} On citation network datasets, we apply a two-layer wGCN model. Specifically, we perform random walk operation 8 times for each node, and stop after passing 4 different nodes every time on Cora dataset. And the decay rate (i.e., $\alpha$, mentioned in section 3.1) is set to 0.8 and the mixing ratio coefficient (i.e., $\lambda$, mentioned in section 3.2) is set to 0.9; on Citation dataset, the random walk operation is performed 3 times for each node, and stop after passing 4 different nodes every time. The decay rate is set to 0.8 and the mixing ratio coefficient is set to 0.73; on Pubmed dataset, the random walk operation is performed 5 times for each node, and also stop after passing 4 different nodes every time. The decay rate and the mixing ratio coefficient ars set to 0.8 and 0.9, respectively. The remaining parameter settings follow the settings in \cite{14}.

\noindent\textbf{Social Network Datasets.} On social network datasets, we apply a three-layer wGCN model. Specifically, the shape of the parameter matrix ($W_1,W_2,W_3$) of the three-layer model, the number $T$ of random walk operation from each node, the decay rate $\alpha$ and the mixing ratio coefficient $\lambda$ on 8 subdatasets are shown in Table 2. In addition, the learning rates on subdatasets of Ego-Facebook and Ego-Gplus are set to 0.02 and 0.01 respectively. And the random walk operation for each node is also stopped after passing 4 different nodes every time.

\vspace{1.5ex}

\subsection{Baselines and Previous Approaches}

\noindent\textbf{Citation Network Datasets.} On citation network datasets (Citeseer, Cora, and Pubmed), our method is compared with the same strong baselines and previous approaches as specified in \cite{14}, including label propagation (LP) \cite{20}, semi-supervised embedding (SemiEmb) \cite{21}, manifold regularization (ManiReg) \cite{22}, iterative classification algorithm (ICA) \cite{23} and Planetoid \cite{24}. Here, DeepWalk is a method based on random walks, as stated at the beginning of the article, whose sampling strategy can be seen as a special case of node2vec with $p=1$ and $q=1$. As for method named GCN, which is the first method to achieve convolution on the graph, it is the best performing baseline method.

\noindent\textbf{Social Network Datasets.} On social network datasets (Ego-Facebook and Ego-Gplus), our method is compared against Deepwalk, GraRep and again GCN, which is the strongest baseline in the above experiment. And here, GraRep \cite{10} works by utilizing the adjacency matrix of each order and defining a more accurate loss function that allows non-linear combinations of different local relationship information to be integrated.

\subsection{Results}

The results of our comparative evaluation experiments are summarized in Tables 3 and 4.

\begin{table}[h] % [h] forces the table to be output where it is defined in the code (it suppresses floating)
	\renewcommand{\tablename}
	\caption{\centering{\textbf{Table 3: The results of classification accuracy on citation network datasets.}}}
	\vspace{2ex}
	\renewcommand{\arraystretch}{1.2}
	\setlength\tabcolsep{16pt}
	\centering
	\begin{tabular}{c c c c}
		\toprule
		\textbf{Method} & \textbf{Cora} & \textbf{Citeseer} & \textbf{Pubmed} \\
		\midrule
		ManiReg & 59.5 & 60.1 & 70.7 \\
		SemiEmb & 59.0 & 60.1 & 71.1 \\
		LP & 68.0 & 45.3 & 63.0 \\
		DeepWalk & 67.2 & 43.2 & 65.3 \\
		ICA & 75.1& 69.1 & 73.9 \\
		Planetoid & 75.7 & 64.7 & 77.2 \\
		GCN & 81.5 & 70.3 & 79.0 \\
		\midrule
		wGCN & \textbf{81.8} & \textbf{72.4} & \textbf{79.5} \\
		\bottomrule
	\end{tabular}
\end{table}

As shown in Table 2, our method achieves the best results on all datasets, and compared with the strongest baseline, our method improve upon GCN by a margin of 0.3, 2.1 and 0.5 on Cora, Citeseer and Pubmed respectively. 

Next, we use the social network datasets (Ego-Facebook and  Ego-Gplus) for the experiments, and compare the experimental results with the classic method based on adjacency matrix $--$ GraRep [] and based on random walk $--$ Deepwalk [], and the best performing baseline method $--$ GCN []. The experimental results are shown in the Table 3:

\begin{table}[h] % [h] forces the table to be output where it is defined in the code (it suppresses floating)
	\renewcommand{\tablename}
	\caption{\centering{\textbf{Table 4: The results of classification accuracy on social network datasets.}}}
	\vspace{2ex}
	\renewcommand{\arraystretch}{1.2}
	\setlength\tabcolsep{6pt}
	\centering
	\begin{tabular}{c c c c c}
		\toprule
		\textbf{Method} & \textbf{107Ego(F)} & \textbf{414Ego(F)} & \textbf{1684Ego(F)} & \textbf{1912Ego(F)} \\
		\midrule
		DeepWalk & 77.5 & 79.2 & 64.4 & 66.5 \\
		GraRep & 90.0 & 85.4 & 76.3 & 77.0 \\
		GCN & 92.5 & 93.8 & 81.9 & 77.0 \\
		\midrule
		wGCN & \textbf{95.0} & \textbf{97.9} & \textbf{85.0} & \textbf{81.0} \\
		\bottomrule
	\end{tabular}

	\vspace{4ex}

	\setlength\tabcolsep{5pt}
	\begin{tabular}{c c c c c}
		\toprule
		\textbf{Method} & \textbf{2106Ego(G)} & \textbf{3600Ego(G)} & \textbf{5249Ego(G)} & \textbf{7461Ego(G)} \\
		\midrule
		DeepWalk & 75.8 & 47.5 & 76.2 & 65.6 \\
		GraRep & 87.9 & 62.3 & 98.9 & 77.4 \\
		GCN & 92.3 & 95.7 & \textbf{99.5} & 80.2 \\
		\midrule
		wGCN & \textbf{93.4} & \textbf{98.1} & \textbf{99.5} & \textbf{83.1} \\
		\bottomrule
	\end{tabular}
\end{table}

As can be seen from Table 3, the experimental results of both parts of our method are significantly higher than the results of Deepwalk and GraRep. And except for '5249Ego(G)' (the result is as good as the result of GCN), the other results of our method are also better than the results of GCN.

\section{Conclusion}

In this paper, we have designed a new framework combined with random walk $--$ wGCN, which can reconstruct the graph to capture higher-order features through random walk, and can effectively aggregate node information. We conduct experiments on a series of datasets (citation network and social network, directed and undirected). The results have shown that wGCN can effectively generate embeddings for nodes of unknown category and get the better results than the baseline methods.

There are many extensions and potential improvements to our method, such as exploring more random walk methods with different strategies and extending wGCN to handle multi-graph mode or time-series-graph mode. Another interesting direction for future work is to extend the method to be able to handle edge features, which will allow the model to have wider applications.

\ifCLASSOPTIONcompsoc
  \section*{Acknowledgments}
\else
  \section*{Acknowledgment}
\fi

This work is supported by the Research and Development Program of China (No.2018AAA0101100), the Fundamental Research Funds for the Central Universities, the International Cooperation Project No.2010DFR00700, Fundamental Research of Civil Aircraft No. MJ-F-2012-04, the Beijing Natural Science Foundation (1192012, Z180005) and National Natural Science Foundation of China (No.62050132).

\ifCLASSOPTIONcaptionsoff
  \newpage
\fi

\bibliographystyle{IEEEtran}
\bibliography{mybibfile}
\end{document}